\documentclass[sigconf]{acmart}
\AtBeginDocument{%
  }
\copyrightyear{2026}
\acmYear{2026}
\setcopyright{cc}
\setcctype{by}
\acmConference[WWW '26]{Proceedings of the ACM Web Conference 2026}{April 13--17, 2026}{Dubai, United Arab Emirates}
\acmBooktitle{Proceedings of the ACM Web Conference 2026 (WWW '26), April 13--17, 2026, Dubai, United Arab Emirates}
\acmPrice{}
\acmDOI{10.1145/3774904.3792934}
\acmISBN{979-8-4007-2307-0/2026/04}

\acmSubmissionID{srp641}

\usepackage[table,xcdraw]{xcolor}
\usepackage[labelfont=bf, format=hang, format=plain]{caption}
\usepackage{color, colortbl}
\usepackage{stfloats}
\usepackage{enumitem}
\usepackage{tabularx}
\usepackage{xstring}
\usepackage{multirow}
\usepackage{xspace}
\usepackage{url}
\usepackage{subcaption}
\usepackage{enumitem}
\usepackage{svg}\usepackage[hang,flushmargin]{footmisc}

\usepackage{siunitx}
\usepackage{pifont}
\newcommand{\xmark}{\ding{55}} % ✗
\newcommand{\cmark}{\ding{51}} % ✓
\usepackage{graphicx}
\usepackage{xcolor}
\usepackage{amsmath}
\usepackage{amsfonts}
\usepackage{bbm}
\usepackage{geometry}
\usepackage{algorithm}
\usepackage{algpseudocode}
\usepackage{amsfonts}
\usepackage{mathtools}
\usepackage{float}
\usepackage[10pt]{moresize}
\usepackage{color}
\usepackage{epsfig}
\usepackage{graphicx}

\usepackage{comment}
\usepackage{pifont}

\usepackage{microtype}
\usepackage[font=smaller]{caption}
\usepackage{tabularx}
\usepackage{adjustbox}
\usepackage{array}
\usepackage{booktabs}
\usepackage{colortbl}
\usepackage{float,wrapfig}
\usepackage{makecell}
\usepackage{hhline}
\usepackage{subcaption} %
\usepackage[percent]{overpic}
\usepackage{graphbox}
\usepackage{diagbox}
\usepackage{stmaryrd}
\usepackage{wrapfig}
\usepackage{tikz}
\usepackage{fontawesome5}
\usepackage[most]{tcolorbox} % Put this in your preamble
\usepackage{ifthen}

\newcommand{\myparagraph}[1]{
\vspace{4pt}\noindent
\textbf{#1.}
}

\usepackage{enumitem}

\usepackage{arydshln}

\usepackage{array}
\newcolumntype{L}[1]{>{\raggedright\let\newline\\\arraybackslash\hspace{0pt}}m{#1}}
\newcolumntype{C}[1]{>{\centering\let\newline\\\arraybackslash\hspace{0pt}}m{#1}}
\newcolumntype{R}[1]{>{\raggedleft\let\newline\\\arraybackslash\hspace{0pt}}m{#1}}
\RequirePackage[font=normalsize,skip=3pt,subrefformat=parens]{subcaption}
\RequirePackage[format=plain,labelformat=simple,labelsep=period,font=small,compatibility=false]{caption}

\colorlet{ochre}{blue!30!yellow!70!}
\definecolor{aquamarine}{rgb}{0.5, 1.0, 0.83}
\begin{document}

\title{NERdME: a Named Entity Recognition Dataset for Indexing Research Artifacts in Code Repositories 
}


\author{Genet Asefa Gesese}
\authornotemark[1]
\orcid{0000-0003-3807-7145}
\affiliation{%
  \institution{FIZ Karlsruhe}
\city{Eggenstein-Leopoldshafen}
\country{Germany}
%    \country{}
}
\affiliation{%
 \institution{Karlsruhe Institute of Technology}
% \country{Germany}
% \country{}
\city{Karlsruhe}
\country{Germany}
}
% \email{genet.gesese@partner.kit.edu}

\author{Zongxiong Chen}
\authornote{These authors contributed equally to this work.}
\orcid{0000-0003-2452-0572}
\affiliation{%
 \institution{Fraunhofer FOKUS}
\city{Berlin}
\country{Germany}
}
% \email{zongxiong.chen@fokus.fraunhofer.de}

\author{Shufan Jiang}
\authornotemark[1]
\orcid{0000-0002-8486-3158}
\affiliation{%
 \institution{IRIT, UT, CNRS, Toulouse INP}
 \city{Toulouse}
\country{France}
}
\affiliation{%
 \institution{Université de Toulouse Jean Jaurès }
 \city{Toulouse}
\country{France}
}

%\settopmatter{authorsperrow=4}

\author{Mary Ann Tan}
\orcid{0000-0003-3634-3550}
\affiliation{%
  \institution{FIZ Karlsruhe}
    \city{Eggenstein-Leopoldshafen}
    \country{Germany}
}
\author{Zhaotai Liu}
\orcid{0009-0004-4740-5790}
\affiliation{%
 \institution{FIZ Karlsruhe}
\city{Eggenstein-Leopoldshafen}
\country{Germany}
}
% \email{ann.tan@fiz-karlsruhe.de}
% \email{uvyrq@student.kit.edu}

\author{Sonja Schimmler}
\orcid{0000-0002-8786-7250}
\affiliation{%
  \institution{Fraunhofer FOKUS \& Technische Universität Berlin}
  \city{Berlin}
  \country{Germany}
}
% \email{sonja.schimmler@fokus.fraunhofer.de}

\author{Harald Sack}
\orcid{0000-0001-7069-9804}
\affiliation{%
 \institution{FIZ Karlsruhe}
\city{Eggenstein-Leopoldshafen}
\country{Germany}
}
\affiliation{%
 \institution{Karlsruhe Institute of Technology}
% \country{Germany}
% \country{}
\city{Karlsruhe}
\country{Germany}
}
% \email{harald.sack@fiz-karlsruhe.de}
\renewcommand{\shortauthors}{Genet Asefa Gesese et al.}
\newcommand{\zong}[1]{\textcolor{blue}{#1}}
\newcommand{\gent}[1]{\textcolor{blue}{#1}}
\newcommand{\sufang}[1]{\textcolor{blue}{#1}}






%%% Local Variables:
%%% mode: LaTeX
%%% TeX-master: "../main.tex"
%%% End:

\begin{abstract}
Existing scholarly information extraction (SIE) datasets focus on scientific papers and overlook implementation-level details in code repositories. README files describe datasets, source code, and other implementation-level artifacts, however, their free-form Markdown offers little semantic structure, making automatic information extraction difficult. To address this gap, NERdME is introduced: 200 manually annotated README files with over \num{10000} labeled spans and 10 entity types. Baseline results using large language models and fine-tuned transformers show clear differences between paper-level and implementation-level entities, indicating the value of extending SIE benchmarks with entity types available in README files. A downstream entity-linking experiment was conducted to demonstrate that entities derived from READMEs can support artifact discovery and metadata integration.
\end{abstract}

\maketitle

%%% Local Variables:
%%% mode: LaTeX
%%% TeX-master: "../main.tex"
%%% End:

\section{Introduction}
Recently, scholarly information extraction (SIE) has gained increasing attention as scientific knowledge broadly spreads across web-based platforms, including digital libraries, data repositories, and open-source code hosting services. While datasets such as SciERC~\cite{luan2018multi}, SciREX~\cite{jain2020scirex}, SciDMT~\cite{pan2024scidmt} and GSAP-NER~\cite{otto2023gsap} have advanced the extraction of entities from scientific articles, these paper-centric datasets overlook a major component of the research ecosystem on the Web: research software repositories. Platforms, such as GitHub, host implementation-level metadata essential for reproducibility, e.g., datasets, dependencies, licenses, and usage details, yet such information rarely appear in the corresponding publications and are instead in README files written in free-form Markdown that provides high-level document structure rather than explicit semantic cues. As a result, the scholarly metadata encoded in READMEs remains inaccessible to automated SIE systems and unlinked within the broader Web of research artifacts.
% \zong{(e.g., OpenAlex, OpenAIRE)}
% \zong{limiting downstream applications such as FAIR metadata pipelines, artifact search, and integration into open science knowledge graphs (e.g., OpenAlex, OpenAIRE)}.

As shown in Table~\ref{tbl:comparison}, existing SIE datasets therefore capture either only paper-level research entities such as tasks, methods, and metrics, or implementation-level information such as software, programming languages, or licenses, none cover entities in both levels comprehensively, which is common in READMEs. Additionally, different document types may refer to the same research entity in different ways: e.g., a scholarly paper might reference a dataset by citing its datasheet, whereas a technical document might refer to the same dataset by linking to its download page. Although Hidden Entity~\cite{gan2025hidden} targets README content, it focuses on URL recognition and classification; without providing span boundaries or annotating entity identifiers inside the URLs, it would not enable the connection of conceptual entities in scientific writing and implementation-related entities in software repositories. To the best of our knowledge, no existing resource provides span-level annotations in README files with the implementation-oriented entities in software repositories.

\begin{table}[tb]
% \setlength{\tabcolsep}{3pt}
% \cmark and \xmark indicates the dataset supports and does not support the capability, respectively.
\caption{Comparison of NERdME with existing SIE datasets.}
\vspace{-0.5cm}
\label{tbl:comparison}
\begin{center}
\footnotesize
\setlength{\tabcolsep}{1.5pt}
% \resizebox{\linewidth}{!}{%
\begin{tabular}{lllcccccc}
\toprule
% Dataset & Source Type & Granularity & \#Docs & Entity Types & Nested  & Focus \\
\multirow{1}{*}{Dataset}   & \multirow{1}{*}{Source} & \multirow{1}{*}{\#ET} & \multirow{1}{*}{\#Doc} &\multirow{1}{*}{\#Mention}  & \multirow{1}{*}{Manual} & \multirow{1}{*}{ILE} & \multirow{1}{*}{PLE} & \multirow{1}{*}{Span} \\  
%                          &           &      Strategy         &       &       &    Types  &  Level Entities   & Level Entities & \\
\midrule
SciERC~\cite{luan2018multi} & Paper abstract & 6 & 500  & \num{8089}           & \cmark & \xmark & \cmark & \cmark \\
SciREX~\cite{jain2020scirex} & Full papers     & 4 & 438  & \num{156931}        & \cmark & \xmark & \cmark & \cmark \\
SciDMT~\cite{pan2024scidmt} & Full papers          & 3 & 48k  & \num{1830886}     & \xmark & \xmark & \cmark & \cmark \\
GSAP-NER~\cite{otto2023gsap} & Full papers          & 10 & 100  & \num{54598}   & \cmark & \xmark & \cmark & \cmark \\
SoMeSci~\cite{schindler2021somesci} & Full papers & 9 & 1367 & \num{7237}   & \cmark & \cmark & \xmark & \cmark \\
Hidden Entity~\cite{gan2025hidden}  & GitHub READMEs & 4 & 811 & \num{1439}     &  \cmark & \cmark & \xmark & \xmark \\
\midrule 
NERdME (Ours)                       & GitHub READMEs  & 10 & 200  & \num{10691} & \cmark & \cmark & \cmark & \cmark \\
\bottomrule
\end{tabular}
% }
% Nested entities refer to annotations where one entity span is fully or partially contained within another — e.g.,
% “[Transformer-based language model [BERT]{\text{Model}}]{\text{System}}” involves a nested span.
\raggedright \footnotesize{ET: Entity Types; Manual: Manual Annotation; ILE: Implementation-Level Entities; PLE: Paper-Level Entities. ILE/PLE = presence of at least one entity type in that category (not necessarily full coverage)}
\end{center}
\vspace{-0.5cm}
\end{table}

% SciERC: Task, Method, Metric, Material, Other-ScientificTerm and Generic
% SciREX: Dataset , Metric , Task , Method
% GSAP-NER: MLModel, ModelArchitecture, MLModelGeneric, Method, Task, Dataset, DatasetGeneric, DataSource, ReferenceLink, URL
% Gan: dataset direct link, dataset landing page ,software, other
% SciDMT: 100 manually annotate, 48,049 machine-learning

% ✗
% ✓

To address this gap, NERdME is introduced as a new named entity recognition (NER) dataset built from GitHub README files, with three main contributions:
1) \textbf{NERdME}. A manually curated dataset of 200 README files annotated with 10 scholarly and technical entity types. To the best of our knowledge, NERdME is the first resource to jointly capture paper-level entities (e.g., {CONFERENCE}) and implementation-level entities (e.g., {SOFTWARE}) in README documentation. The dataset and annotation guidelines are publicly available on Zenodo\footnote{\url{https://zenodo.org/records/17571420}}. 
2) \textbf{Benchmarking.} A state-of-the-art (SOTA) evaluation using zero-shot LLMs and fine-tuned transformers. NERdME extends existing SIE benchmarks with additional entity types, and the results highlight limitations of SOTA approaches in recognizing rare and fine-grained entities such as WORKSHOP and ONTOLOGY.
3) \textbf{Downstream Applicability.} The dataset’s downstream use is demonstrated through an entity-linking experiment that aligns extracted dataset mentions with Zenodo records, showing its value for artifact discovery and metadata integration.
% \input{sections/02-related-work}
% %%% Local Variables:
% %%% mode: LaTeX
% %%% TeX-master: "../main.tex"
% %%% End:

\section{Dataset Construction}
\myparagraph{Dataset Collection \& Annotation Tag Selection}\label{sub:data-collection}
We compiled 200 GitHub README files accompanying data science papers from Papers with Code (PwC)
, selected from an intial pool of over \num{2600} candidate repositories for rich scholarly context (e.g., references to papers, datasets, software) and coverage of our target entity categories . 
{The dataset size reflects the cost of high-quality manual annotation: each README was annotated by three annotators and curated by an expert, requiring approximately one hour of human effort per file, making 200 READMEs a practical balance between scale and reliability.}
The annotation tag set comprises ten entity types from the NFDI4DS ontology~\cite{gesese2024nfdi4dso}: {CONFERENCE}, {DATASET}, {EVALUATION METRIC}, {LICENSE}, {ONTOLOGY}, {PROGRAMMING LANGUAGE}, {PROJECT}, {PUBLICATION}, {SOFTWARE}, and {WORKSHOP}. 
These types cover both paper-level concepts (CONFERENCE, DATASET, WORKSHOP, PUBLICATION, EVALUATION METRIC) and implementation-level concepts (SOFTWARE, PROGRAMMING LANGUAGE, LICENSE), which were studied only separately~\cite{schindler2021somesci,jain2020scirex}. 
Definitions of these entity types can be found in the official NFDI4DS documentation~\footnote{\url{https://ise-fizkarlsruhe.github.io/NFDI4DS-Ontology/}}, and illustrative examples are provided in the annotation guidelines. To avoid missing rare entity types during annotation, a two-stage selection strategy was employed. First, 150 README files were manually selected, ensuring that each file contained at least one of the target entity types and that all entity types were represented in a minimum of 15 files. Second, the remaining 50 README files were randomly chosen from the rest of the repositories to ensure diversity and minimize selection bias.

\begin{figure}[tb]
    % \vspace{-0.3cm}
    \centering
    \includegraphics[width=\linewidth]{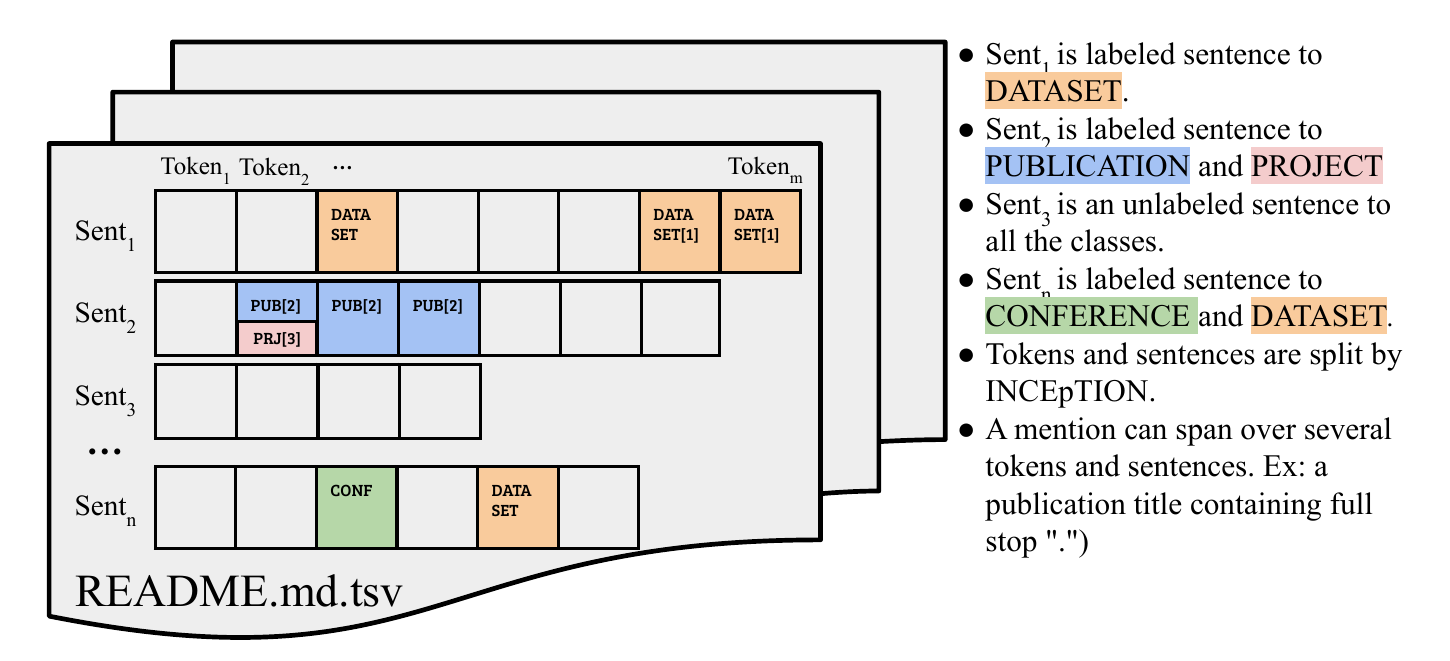}   
   \vspace{-0.7cm}
    \caption{Illustration of an annotated README file in the dataset. Note that unlabeled sentences are NOT negative samples.}
    \label{fig:dataset_illu}
   \vspace{-0.5cm}
\end{figure}

\myparagraph{Annotation and Curation} 
Each selected README was annotated by three independent annotators with a background in computer science, using INCEpTION~\footnote{\url{https://inception-project.github.io/}}. 
All annotators received training and followed detailed annotation guidelines to ensure consistent interpretation of entity definitions. Entity mentions are considered across the entire README file, including descriptive paragraphs, embedded code snippets, citation blocks, and URLs. Entity spans can be nested (e.g., a SOFTWARE name within a PUBLICATION title), and may not always be separated by spaces or punctuation (e.g., a DATASET name embedded in a URL). To ensure consistency and unambiguous evaluation, two curation rules were applied. First, only spans annotated by at least two annotators were retained, ensuring high-confidence labels. Second, sentence-level consistency was enforced for each entity type: if a sentence contained both valid (agreed-upon) and filtered (disagreed) spans for the same entity type, all spans of that type in the sentence were removed. 
% \zong{This conservative policy prevents partially annotated examples from misleading models during training. -> repeated}
For example, consider the following illustrative sentence: \textit{“The experiments conducted to evaluate the NERdME dataset are based on Python.”} Two annotators agreed on the {DATASET} span but disagreed on whether “Python” should be labeled as {SOFTWARE} or {PROGRAMMING LANGUAGE}. In this case, both {SOFTWARE} and {PROGRAMMING LANGUAGE} spans would be removed, while the {DATASET} span is retained. This conservative strategy enforces type-level consistency within each sentence while retaining as many labeled sentences as possible. Figure~\ref{fig:dataset_illu} illustrates an example of an annotated README segment, where both labeled and unlabeled sentences are present. 

% link to edit the figure: https://docs.google.com/drawings/d/1PIrZfZQECabKF6l9sDQHcqSNKQPxnan3rJmiN-3QTUg/edit?usp=sharing
\myparagraph{Inter-Annotator Agreement}
Annotation consistency was measured using Krippendorff’s alpha~\cite{Krippendorff2011ComputingKA}, yielding a score of $0.70 \pm 0.19$ across all README files. This indicates substantial agreement among annotators and supports the reliability of the dataset for training and evaluating entity extraction models.

\begin{figure}[tb]
    \centering
    \includegraphics[
        width=0.90\linewidth,
        % trim={0cm 0.23cm 0cm 0.2cm}, % left bottom right top
        % clip
    ]{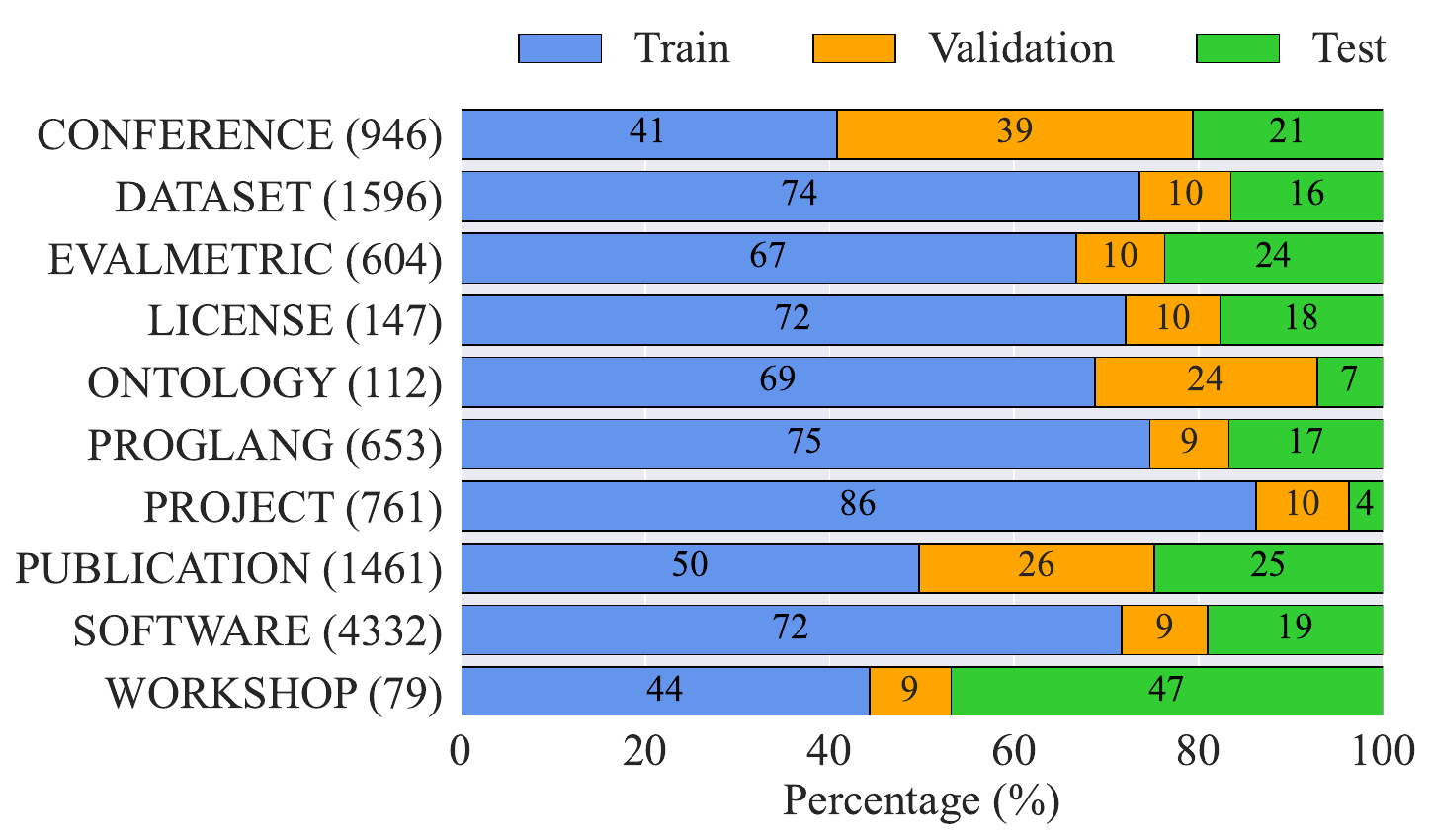}
     \vspace{-0.5cm}
    \caption{Span statistics in NERdME. Bars show the proportion of spans across train, validation, and test splits for each entity type; numbers in parentheses indicate total spans per type.}
    \label{fig:nerdme}
    \vspace{-0.43cm}
\end{figure}

\myparagraph{Dataset Statistics}
The NERdME stored in WebAnno TSV format consists of 200 README files with \num{10691} annotated entity spans, of which \num{4328} are unique spans. The dataset is divided by document into training (70\%), validation (10\%), and test (20\%) splits. Since entity types are not uniformly distributed across files, the proportion of each entity type within a split differ from the overall ratio (cf. Figure~\ref{fig:nerdme}). For example, {CONFERENCE} spans account for 41\%, 39\%, and 21\% of the training, validation, and test sets, respectively. 
Labeled span counts vary considerably across entity types; common entities such as SOFTWARE and DATASET appear frequently in READMEs, whereas more specialized categories like WORKSHOP and ONTOLOGY occur far less often, reflecting their naturally limited presence in project documentation.
We preserve this natural distribution to allow models to learn from the imbalances encountered in real-world settings.

\myparagraph{Linguistic Statistics of Entity Groups} We follow the paper and implementation level concepts introduced previously, and analyze four intrinsic linguistic properties, i.e., \textbf{uniqueness} (ratio of unique mentions)~\cite{malvern2012measures}, \textbf{length} (characters per mention)~\cite{baayen2001word}, \textbf{fluency} (language-model perplexity from DistilGPT-2)
%~\cite{sanh2019distilbert}
, \textbf{variability} (normalized edit distance by Levenshtein)~\cite{lcvenshtcin1966binary}, reporting median values to mitigate the impact of long-tailed span lengths and irregular surface-form distributions. 
As shown in Table~\ref{tbl:comp-paper-impl}, paper-level entities show substantially higher lexical uniqueness (UniqueRatio: 0.56 vs 0.29) and longer surface forms (FormLength: 10.00 vs 6.00), reflecting more descriptive naming conventions such as publication titles or dataset names.
In contrast, implementation-level entities are shorter and more standardized, frequently reusing common labels for software, programming language, and licenses. Their symbol-heavy structure leads to noticeably higher language-model perplexity (PPL: \num{2509} vs \num{1326}), and they also exhibit slightly greater surface-form variation (EditDist: 0.94 vs 0.92) due to versioned and noncanonical naming patterns. These differences highlight the distinct expression styles of paper- and implementation-level entities, underscoring the need for a dataset that spans both.

\begin{table}[t]
\caption{Linguistic comparison of entity groups. }
\vspace{-0.3cm}
\label{tbl:comp-paper-impl}
\begin{center}
\footnotesize
\setlength{\tabcolsep}{1.5pt}
% \resizebox{\linewidth}{!}{%
\begin{tabular}{cccccc}
\toprule
\multirow{1}{*}{Group} & \multirow{1}{*}{UniqueRatio} & \multirow{1}{*}{FormLength} & \multirow{1}{*}{PPL} &\multirow{1}{*}{EditDist} & \\  
\midrule
Paper           & 0.56 & 10.00  & \num{1326} & 0.92 \\
Implementation  & 0.29 &  6.00  & \num{2509} & 0.94 \\
\bottomrule
\end{tabular}
% }
\end{center}
\raggedright\footnotesize *All values are medians computed per group, covering uniqueness, form length, language-model perplexity, and edit distance, to reduce the influence of outliers.
\vspace{-0.5cm}
\end{table}

% \input{sections/04-baseline-models}
 %%% Local Variables:
%%% mode: LaTeX
%%% TeX-master: "../main.tex"
%%% End:
% \vspace{-3mm}

% \vspace{-0.3cm}
\section{Experiments}

\subsection{NER Task}

\myparagraph{Experimental Setup}
The NER task on the NERdME dataset is evaluated using 1) zero-shot LLMs (e.g., Mistral-7B-Instruct-v0.3, LLaMA3.1-8B, GPT-4o-mini, DeepSeek-Chat, and Gemini-2.0-Flash) and 2) fine-tuned transformers for token classification (e.g., SciBERT and RoBERTa-base) as baselines~\footnote{All prompts, model configurations, and training details are included in the \url{https://github.com/chenzongxiong/nerdme}.}. 
For zero-shot LLMs, we annotate each sentence once and filter predicted spans by entity type during evaluation to obtain type-specific scores. LLMs are prompted with temperature equal to 0 for deterministic decoding.
{For fine-tuned transformers, we train a token classifier per entity type using only its corresponding labeled sentences}, and each model predicts only that target type at test time. Out-of-box transformers are fine-tuned using a standard BIO tagging scheme with a learning rate $5e^{-5}$, batch size 16, and 10 epochs.
% {All prompts, model configurations, and training details are included in the GitHub repository
% ~\footnote{\footnotesize \url{https://github.com/chenzongxiong/nerdme}}
% to ensure reproducibility.}
% {In the GitHub repository~\footnote{\footnotesize \url{https://github.com/chenzongxiong/nerdme}}
% , All prompts, model configurations, and training details are included
% to ensure reproducibility.}
This setup allows us to examine NERdME’s capacity to reveal performance gaps between general-purpose and task-adapted models, assess bias in LLMs, and validate the dataset’s applicability for training effective domain-specific NER systems.

\myparagraph{Evaluation Setup}
% Entity extraction is evaluated at the span level for each entity type independently, in order to prevent frequent classes, such as SOFTWARE, from dominating model behavior and to enable a fair, type-specific comparison between filtered zero-shot LLM predictions and single-type fine-tuned models.
{Since NERdME provides type-selective annotations, jointly scoring all labels would incorrectly treat unlabeled spans as negatives, producing unreliable precision/recall. We therefore evaluate each entity type independently to ensure valid, type-specific comparison between filtered zero-shot LLM predictions and single-type fine-tuned models.}
This design avoids cross-type interference and isolates the intrinsic extractability of each entity type. Extraction quality is measured at the character level to avoid tokenization mismatches, and F1 scores are reported under two criteria: Exact Match (boundaries must match exactly) and Partial Match (any character overlap counts as correct). %For each entity type, F1 scores are reported under two span-matching criteria: Exact Match, where the predicted span must match the gold boundaries exactly, and Partial Match, where any character overlap with the gold span is considered correct.

\myparagraph{Results}
% \vspace{2}
% \noindent\vspace{2pt}\textbf{Results.}
% \input{misc/exp-entity-wise2}
\begin{table*}[hbt!]

\caption{Best results of various language models for individual entity type on NERdME under partial and exact match F1 score. F1-P. and F1-E. represent partial and exact match F1 score, respectively.  The complete scores for all models and entity types are available in 
GitHub repository.
%our GitHub repository ~\protect\footnotemark.
}
\label{tbl:exp-classwise}
\vspace{-0.3cm}
\centering
\resizebox{\linewidth}{!}{%
\footnotesize
\begin{tabular}{cccccccccccc}
\toprule
& Approach & CONFERENCE & DATASET & EVALMETRIC & LICENSE & ONTOLOGY & PROGLANG & PROJECT & PUBLICATION & SOFTWARE & WORKSHOP \\
%& \#Span & 946 & 1596 & 604 & 147 & 112 & 653 & 761 & 1461 & 4332 & 791 \\
%& \#Span$_{test}$ & 195 & 263 & 143 & 26 & 8 & 109 & 28 & 362 & 824 & 37 \\

\cmidrule(lr){1-12}
\multirow{2}{*}{F1-P.}
        &Zero-shot LLM & 77.38 & 50.98 & 37.07 & \textbf{87.34} & 45.98 & 34.41 & \textbf{56.23} & 70.42 & 29.96 & \textbf{20.68} \\ \cmidrule(lr){2-12}
        &Fine-tuned & \textbf{82.29} & \textbf{69.94} & \textbf{60.18} & 86.62 & \textbf{79.43 }& \textbf{82.72} & 47.36 & \textbf{90.16} & \textbf{77.25} & 13.27 \\ 
\midrule
\multirow{2}{*}{F1-E.}
        &Zero-shot LLM & 47.63 & 45.30 & 33.94 & \textbf{86.15} & 18.52 & 31.69 & \textbf{53.72} & 64.95 & 23.12 & \textbf{20.68} \\ \cmidrule(lr){2-12}

        &Fine-tuned & \textbf{58.38} & \textbf{63.94} & \textbf{44.78} & 78.38 & \textbf{73.13} & \textbf{79.96} & 34.22 & \textbf{82.12} & \textbf{72.46} & 0.00 \\
\bottomrule
\end{tabular}%
}
\vspace{-0.3cm}
\end{table*}

Table~\ref{tbl:exp-classwise} reports the best partial and exact match F1-scores. The results show consistent trends across both zero-shot LLMs and fine-tuned transformers, revealing the following insights.

\textbf{NERdME provides strong supervision signals that enable effective learning accross entity types.}  Entity types with substantial annotated spans show large improvements in exact match F1 when trained in a supervised setting compared to zero-shot LLMs, such as SOFTWARE (\num{4332} spans, 23.12\% $\rightarrow$ 72.46\%), DATASET (\num{1596} spans, 45.30\% $\rightarrow$ 63.94\%), PUBLICATION (\num{1461} spans, 64.95\% $\rightarrow$  82.12\%), and EVALMETRIC (\num{604} spans, 33.94\% $\rightarrow$  44.78\%). These gains arise because fine-tuned models directly learn from NERdME’s span-level annotations, whereas zero-shot LLMs have no exposure to the dataset’s domain-specific naming patterns or boundary structures. 
% Moreover, in zero-shot settings, paper-level entities tend to achieve higher F1 because their terminology is more standardized and frequent in scholarly text, which LLMs are likely to have encountered during pretraining. In contrast, implementation-level entities are more diverse in naming and formatting, making them harder to identify without supervision
%, though LICENSE shows relatively higher zero-shot performance, possibly due to more standardized naming conventions
% . 
Fine-tuned models, however, substantially improves performance for most entity types in both groups, highlighting that NERdME provides coherent and learnable supervision for span-level generalization.

\textbf{Long-tailed entity distribution shapes extractability across entity types.}
Common entity classes such as SOFTWARE (\num{4332} spans) and DATASET (\num{1596} spans) achieve consistently higher F1 scores, while long-tail classes such as WORKSHOP (79 spans) and ONTOLOGY (112 spans) show lower performance. This long-tailed distribution reflects the imbalance of README documentation and demonstrates that NERdME provides a realistic testbed for assessing model behavior across both common and long-tail entities.

\textbf{Span-level complexity exposes boundary alignment challenges.}
Across almost all entity types, models exhibit large drops from partial to exact match F1 score, for example, CONFERENCE decreases from 77.38\% to 47.63\% and 82.29\% to 58.38\% in zero-shot LLMs and fine-tuned models, respectively. Since evaluation is conducted per entity type independently, this gap reflects the difficulty in precise span boundaries rather than type confusion. This indicates that NERdME includes many entities with flexible or context-dependent boundaries, making it a strong benchmark for assessing fine-grained span localization, which is essential for SIE tasks.

% \vspace{-0.3cm}
\subsection{Downstream Task: Entity Linking (EL)}
While NERdME is designed for NER, its annotated entities also support downstream tasks such as linking to external scholarly resources. To illustrate this, an EL experiment was conducted to align DATASET mentions in NERdME to their corresponding canonical records in Zenodo, demonstrating that the annotated spans can support tasks such as artifact discovery and metadata integration.

% DATASET in NERdME: Unique 661
\myparagraph{Experimental setup}
% All DATASET entities from NERdME were extracted.
To ensure realistic candidate overlap, Zenodo was queried using the surface form each DATASET mention in NERdME, yielding \num{1115} candidate records.
% \zong{Since Zenodo titles are noisy, we manually curated clean gold titles and used GPT-5 as an auxiliary annotator to normalize and assign these titles to NERdME mentions.}
Two lightweight linking strategies were evaluated: 1) \textbf{fuzzy matching}, using token-set ratio scoring, captures surface-level lexical overlap between NERdME and Zenodo titles. 2) \textbf{semantic similarity matching} based on the pretrained all-MiniLM-L6-v2 encoder~\cite{wang2020minilm}.

\myparagraph{Evaluation setup}
Linking performance was measured with ranking and classification metrics. For ranking, we report mean reciprocal rank (MRR), which measures how high the correct Zenodo record appears in the ranked list, and Hits@1/3, capturing the proportion of mentions whose correct match is ranked in the top 1 or top 3 position(s). For classification, we treat linking as a binary decision using the top-ranked candidate and compute Precision (P), Recall (R), and F1 based on whether that candidate matches the gold record.
% \vspace{-0.8cm}
\begin{table}[hb]
\centering
\caption{Entity linking performance on NERdME entities under different linking methods. Best results are marked in bold.}\label{tbl:entity-linking}
\vspace{-0.3cm}
% \resizebox{\linewidth}{!} {
\footnotesize
\begin{tabular}{clcccccc}
\toprule
% \multirow{1}{*}{\textbf{Platform}} & 
\textbf{Method} & \textbf{P} & \textbf{R} & \textbf{F1} & \textbf{Hits@1} & \textbf{Hits@3} & \textbf{MRR} \\
\midrule
% \multirow{2}{*}{Zenodo}  & 
Fuzzy & 26.84 & 13.17 & 17.67 & 15.24 & 15.29 & 15.27 \\
% & 
Semantic & \textbf{65.18} & \textbf{38.68} & \textbf{47.75} & \textbf{37.68} & \textbf{37.72} & \textbf{37.69} \\
% & \zong{TODO: Semantic + FT} & \textbf{65.18} & \textbf{38.68} & \textbf{47.75} & \textbf{37.68} & \textbf{37.72} & \textbf{37.69} \\
% \midrule
% \multirow{2}{*}{PwC} 
% & Fuzzy & 73.63 & 34.90 & 47.35 & 27.50 & 30.00 & 29.70 \\
% & Semantic & \textbf{77.76} & \textbf{85.25} & \textbf{80.21} & \textbf{47.81} & \textbf{50.63} & \textbf{50.26} \\
% & \zong{Semantic + FT} & \textbf{80.86} & \textbf{92.35} & \textbf{86.22} & \textbf{53.75} & \textbf{57.50} & \textbf{55.75} \\

\bottomrule
\end{tabular}
% }
% PwC: test: 320
% PwC: train: 1277
% PwC: total = 320 + 1277 = 1596
\vspace{-0.5cm}
\end{table}

% \begin{table}[!htb]
% \centering
% \caption{Entity Linking performance on NERdME entities under different linking methods.}\label{tbl:entity-linking}
% % \resizebox{\linewidth}{!} {
% \begin{tabular}{lccc}
% \toprule
% \textbf{Method} & \textbf{Precision} & \textbf{Recall} & \textbf{F1}  \\
% \midrule
% Keyword-based & 26.90 & 13.23 & 17.73 \\
% Semantic-based & \textbf{65.60} & \textbf{38.40} & \textbf{48.44} \\
% \bottomrule
% \end{tabular}
% % }
% \end{table}

\myparagraph{Results}
The results show that NERdME’s entity spans are semantically rich enough to support reliable alignment with external scholarly resources, with semantic similarity outperforming fuzzy matching across all metrics (Table \ref{tbl:entity-linking}). It achieves substantially higher F1 (47.75\% vs 17.67\%), Precision (P), and Recall (R), indicating that NERdME mentions contain semantic cues strong enough for accurate linking. Ranking quality also improves, with higher MRR (37.69\% vs 15.27\%) and Hit@1/3 (37.68\% and 37.72\% vs 15.24\% and 15.29\%). These results show that NERdME spans carry lexical and contextual information sufficient for reliable alignment with external scholarly resources, supporting downstream tasks such as artifact discovery, resource indexing, and metadata enrichment.
% Extending linking to other entity types is left for future work.
% \zong{TODO: Add future work, focusing contribution to WWW-community.}

%%% Local Variables:
%%% mode: LaTeX
%%% TeX-master: "../main.tex"
%%% End:

\section{Conclusion}
This paper introduces NERdME, the first NER dataset with over 10,000 labeled spans covering both paper-level and implementation-level scholarly entities from software README files. Extensive evaluations on NER and EL tasks show how NERdME’s diverse entity types and span characteristics affect extraction performance and enable metadata enrichment. Beyond benchmarking, NERdME opens up potential research directions, including scholarly entity coreference resolution, SIE from heterogeneous research artifacts, benchmarking with incomplete data, and structure-aware adaptation of LLMs for domain-specific information extraction.

% \newpage
% \appendix
% \input{sections/07-appendix}
% \vspace{-0.3cm}
\begin{acks}% 
This work was funded by the Federal Ministry of Research, Technology and Space (BMFTR; M532701) / (DFG - NFDI4DS 460234259). %The authors also acknowledge the annotators involved in this work.

\end{acks}
\bibliographystyle{ACM-Reference-Format}
\bibliography{biblio}

\end{document}